# End-to-End Vision-Based Adaptive Cruise Control (ACC) Using Deep Reinforcement Learning

**Zhensong Wei, Corresponding Author**
College of Engineering - Center for Environmental Research and Technology
University of California, Riverside
1084 Columbia Ave, Riverside, CA 92507, USA
E-mail: zwei030@ucr.edu

**Yu Jiang**
College of Engineering - Center for Environmental Research and Technology
University of California, Riverside
1084 Columbia Ave, Riverside, CA 92507, USA
E-mail: yjian091@ucr.edu

**Xishun Liao**
College of Engineering - Center for Environmental Research and Technology
University of California, Riverside
1084 Columbia Ave, Riverside, CA 92507, USA
E-mail: xliao016@ucr.edu

**Xuewei Qi, Ph.D.**
Independent AI Researcher
37066 Edmanton Dr, Sterling Heights, Michigan, 48312
E-mail: qixuewei@gmail.com

**Ziran Wang, Ph.D.**
College of Engineering - Center for Environmental Research and Technology
University of California, Riverside
1084 Columbia Ave, Riverside, CA 92507, USA
E-mail: zwang050@ucr.edu

**Guoyuan Wu, Ph.D.**
College of Engineering - Center for Environmental Research and Technology
University of California, Riverside
1084 Columbia Ave, Riverside, CA 92507, USA
E-mail: gywu@cert.ucr.edu

**Peng Hao, Ph.D.**
College of Engineering - Center for Environmental Research and Technology
University of California, Riverside
1084 Columbia Ave, Riverside, CA 92507, USA
E-mail: haop@cert.ucr.edu

**Matthew Barth, Ph.D.**
College of Engineering - Center for Environmental Research and Technology
University of California, Riverside
1084 Columbia Ave, Riverside, CA 92507, USA
E-mail: barth@cert.ucr.edu

Word Count: 4,964 words + 4 tables = 5,964 words

*Submitted [August 1, 2019]*

**ABSTRACT**

This paper presented a deep reinforcement learning method named Double Deep Q-networks to design an end-to-end vision-based adaptive cruise control (ACC) system. A simulation environment of a highway scene was set up in Unity, which is a game engine that provided both physical models of vehicles and feature data for training and testing. Well-designed reward functions associated with the following distance and throttle/brake force were implemented in the reinforcement learning model for both internal combustion engine (ICE) vehicles and electric vehicles (EV) to perform adaptive cruise control. The gap statistics and total energy consumption are evaluated for different vehicle types to explore the relationship between reward functions and powertrain characteristics. Compared with the traditional radar-based ACC systems or human-in-the-loop simulation, the proposed vision-based ACC system can generate either a better gap regulated trajectory or a smoother speed trajectory depending on the preset reward function. The proposed system can be well adaptive to different speed trajectories of the preceding vehicle and operated in real-time.

**INTRODUCTION**

By leveraging on-board sensors such as radars, the adaptive cruise control (ACC) system is able to automatically adjust the vehicle longitudinal speed to maintain a safe distance with the preceding vehicle and to resume the preset cruising speed when no predecessor is detected. Due to its improved safety performance and driving convenience, the ACC system has been implemented in various models by an increasing number of car manufacturers. Nevertheless, no commercial ACC system has incorporated environmental sustainability within its design loop, despite the fact that the transportation-related activities account for more than 29% of nationwide energy consumption and greenhouse gas (GHG) emissions in the United States (1). With the increasing amount of transportation energy consumption, the development of electric vehicles (EV) appears to be a good solution to this problem. As investigated by (2), the U.S. EV sales in 2018 have increased by 81% compared to those in 2017, reaching around 360k vehicles for the entire year. The U.S. market share of EV sales has been rapidly increasing since 2015, amounting to around 1.2% of total market share in 2018, and is estimated to reach around 8% in 2026 (3). With the huge market share of internal combustion engine (ICE) vehicles and increasing demand of EVs, this paper proposes an end-to-end solution to a vision-based ACC system and evaluates the energy consumption of the proposed system for both types of vehicles.

Among all the types of sensors used in ACC systems, radars and lasers are most common, but they do not necessarily provide richer information than the camera under good weather and lighting conditions. Therefore, camera-based systems that have a stronger capability of sensing the traffic conditions are worth exploring. The conventional method of camera-based ACC or vision-based ACC is to extract depth information from the parallax using multiple front-facing cameras and calculate the optimal cruising speed according to the distance measurement. However, such system usually consists of multiple components and each of the components needs to be optimized separately with tremendous efforts. End-to-end learning, which has been successfully applied to other vehicle automation fields (4-6), can solve this problem by optimizing the entire system using one loss function. Therefore, the system can be trained based on a single principle in a data-driven manner, which largely reduces the effort of human expert.

Not only the end-to-end training has been proved to be effective, the deep reinforcement learning (DRL) has also been widely used in the research of driving strategies. Okuyama et al. proposed a simplified autonomous driving system trained by Deep Q Network (DQN) based on given input images of the street captured by the front-facing car camera (7). Lillicrap et al. adapted the ideas of Deep Q-Learning to the continuous action domain and a car driving scenario was tested for learning the end-to-end policies from raw pixel inputs (8). Sallab et al. proposed a framework for autonomous driving using DRL and tested the system in a simulation environment where complex road curvatures were included (9). Despite the various DRL applications, there was not an end-to-end vision-based ACC system using DRL. Therefore, a deep reinforcement learning (DRL) algorithm is applied in our study to seek an end-to-end solution.

The proposed system is trained and validated in a simulation environment where physical models are built, and feature data, such as vehicle speed, following distance and throttle and braking force, are extracted as system inputs. The system can also achieve a stricter or more relaxed following compared to the traditional ACC algorithm or human-in-the-loop simulation, based on different preset reward functions during the training. During the testing, only the sequence of front-facing camera images and its own speed are required for the system input.

The rest of the paper is organized as follow. Section II discusses the recent work related to ACC and energy consumption modeling. Section III illustrates the workflow of the system, experiment setup, data preprocessing method and the applied deep reinforcement learning network structure. Section IV compares the distance keeping ability of the proposed method to the traditional ACC algorithm and human-in-the-loop simulation in the Unity (10) environment with an evaluation of the energy consumption rate. Section V concludes the paper with some discussions about future work.

**RELATED WORKS**
**ACC Algorithms**
ACC has become a popular technology and has been implemented in many intelligent vehicles in recent years. Many traditional ACC systems are formulated as a bi-level system architecture, where the higher level controller determines the reference acceleration according to the external traffic environment, while the lower level controller track the reference acceleration by controlling the throttle and brake (11). In the research of upper level controller, Gao and Yan proposed an ACC algorithm with the compensation of the preceding vehicle acceleration to follow the preceding vehicle with a stable and safe speed (12). Luo proposed a human-like ACC algorithm using model predictive control (MPC), where safety, comfort and time optimal are considered (13). Shi and Wu proposed an ACC algorithm based on target recognition to follow the vehicles on the curve (14). In (15), the authors proposed a second-order consensus ACC algorithm, which controlled the longitudinal speed of the host vehicle by giving a suggested acceleration, based on the speed difference and gap between the host vehicle and the preceding vehicle. The suggested acceleration is given by:

$$a_{ref} = \beta \cdot (d_{gap} - d_{ref}) + \gamma \cdot (v_{pre} - v_{host}) \qquad (1)$$

where $\beta$ and $\gamma$ are damping gains; $d_{ref}$ is the desired inter-vehicle gap, which is calculated as:

$$d_{ref} = \max(d_{gap}, d_{safe}) \qquad (2)$$

where $d_{safe}$ is the minimum safe distance between two vehicles to avoid collision, and $d_{gap}$ is the time gap-based inter-vehicle distance that is the product of desired time gap and the current speed of host vehicle, which is given by:

$$d_{gap} = v_{host} \cdot t_{gap} \qquad (3)$$

In high-level congestion traffic, the host vehicle may have to stop. In that condition, when the $v_{host}$ approaches to 0, the $d_{gap}$ will be 0, which may cause collision. Thus, a safety threshold, $d_{safe}$, is set up to ensure the host vehicle keep a minimum distance to the preceding vehicle to prevent the collision. In this study, we used this ACC algorithm as the baseline for comparative studies of the proposed ACC system in terms of following distance and energy consumption.

Among the studies for vision-based ACC, Stein et al. used the scale change of the target vehicle in sequentially captured images to determine the bound of range and range rate, which were used as inputs to the proposed system (16). Kanjee et al. used a mono camera to obtain the clearance distance between the preceding and host vehicles using pattern matching (17). Hofmann et al. fused the radar-based ACC and visual perception to design a hybrid ACC system on the highway (18). All the aforementioned vision-based ACC systems required massive image processing techniques or complex road models for extracting the information to the system and therefore hard to adapt the methods to new scenarios. Our proposed system solves this problem by using an end-to-end solution.

**Energy Consumption Models**

Various energy consumption models have been developed by researchers for different vehicle types and environments. For example, Duell et al. proposed an energy consumption estimation model for ICE vehicles using dynamic traffic assignment vehicle trajectories (19). Shibata and Nakagawa developed a mathematical model to calculate the EVs' power consumption for both traveling and air-conditioning (20).

In this work, we considered two types of vehicles, i.e., ICE vehicles and EVs, and used corresponding microscopic emission/energy models for system evaluation. More specifically, we applied the Comprehensive Modal Emissions Model (CMEM) (21) to ICE vehicles, and a hybrid energy consumption rate model (22) for EVs, respectively. These two models are briefly introduced in the following.

*2.4.1 CMEM*

CMEM is designed to estimate the individual ICE vehicle's fuel consumption and emissions.

$$\frac{dFuel}{dt} \approx \lambda \left( k \cdot N \cdot D + \frac{P_{engine}}{\eta_{engine}} \right) \tag{4}$$

Based on the fuel consumption and engine power, the emission can be computed as:

$$\text{Emission} = \frac{dFuel}{dt} \cdot \frac{g_{emission}}{g_{fuel}} \cdot CPF \tag{5}$$

where $\frac{g_{emission}}{g_{fuel}}$ is ratio of the either engine-out emission index and fuel consumption. CPF is the catalyst pass fraction, which is defined as the ratio of tailpipe to engine-out emissions.

*2.4.2 Hybrid Energy Consumption Rate Estimation for EV*

In our study, we implement an EV energy consumption rate estimation model (Type II) adapted from (21) for evaluation. In the equation, the road grade is set to 0 since our simulation environment was built with zero road grade. The equation of the EV energy consumption rate estimation model is shown in **Equation 6**:

$$P_{est} = l_0 + l_1 v + l_2 v^3 + l_3 va + l_4 v^2 + l_5 v^4 + l_6 v^2 a + l_7 v^3 a + l_8 va^2 \tag{6}$$

where $v$ is the speed; $a$ is the acceleration; and $l_i$'s is the associated coefficient of the *i*-th term. Please refer to Table 4 in (21) for the values of associated coefficients.

**METHODOLOGY**

In this section, we present the detailed description of the proposed end-to-end vision-based ACC system where the Double Deep Q Network (DDQN) algorithm and network architecture are adopted.

**System Workflow**

As shown in **Figure 1**, the system consists of two stages, training and testing. During the training, the interaction between the simulation environment and RL network occurs at every time step. The throttle/brake force applied to the system in the previous time step and the following distance between the preceding vehicle and host vehicle in the current time step are measured to calculate the reward function of the RL model. The host vehicle speed sequences and the front camera image sequences are the inputs sent to the RL network for training. Then, the simulation environment is frozen and waits for the throttle/brake force output from the reinforcement learning model. The output force is used to control the vehicle behavior in the next time step.

During the testing, at each time step, the host vehicle speed and the front camera image sequences are sent to the trained DQN model to get the optimal throttle/brake force output. The inference and transmission time is short enough to ensure the real-time working capability of the system. After the entire testing process, the gap statistics and energy consumption can be calculated using the speed trajectory of the vehicle and compared with that of the traditional ACC algorithm and human-in-the-loop simulation.

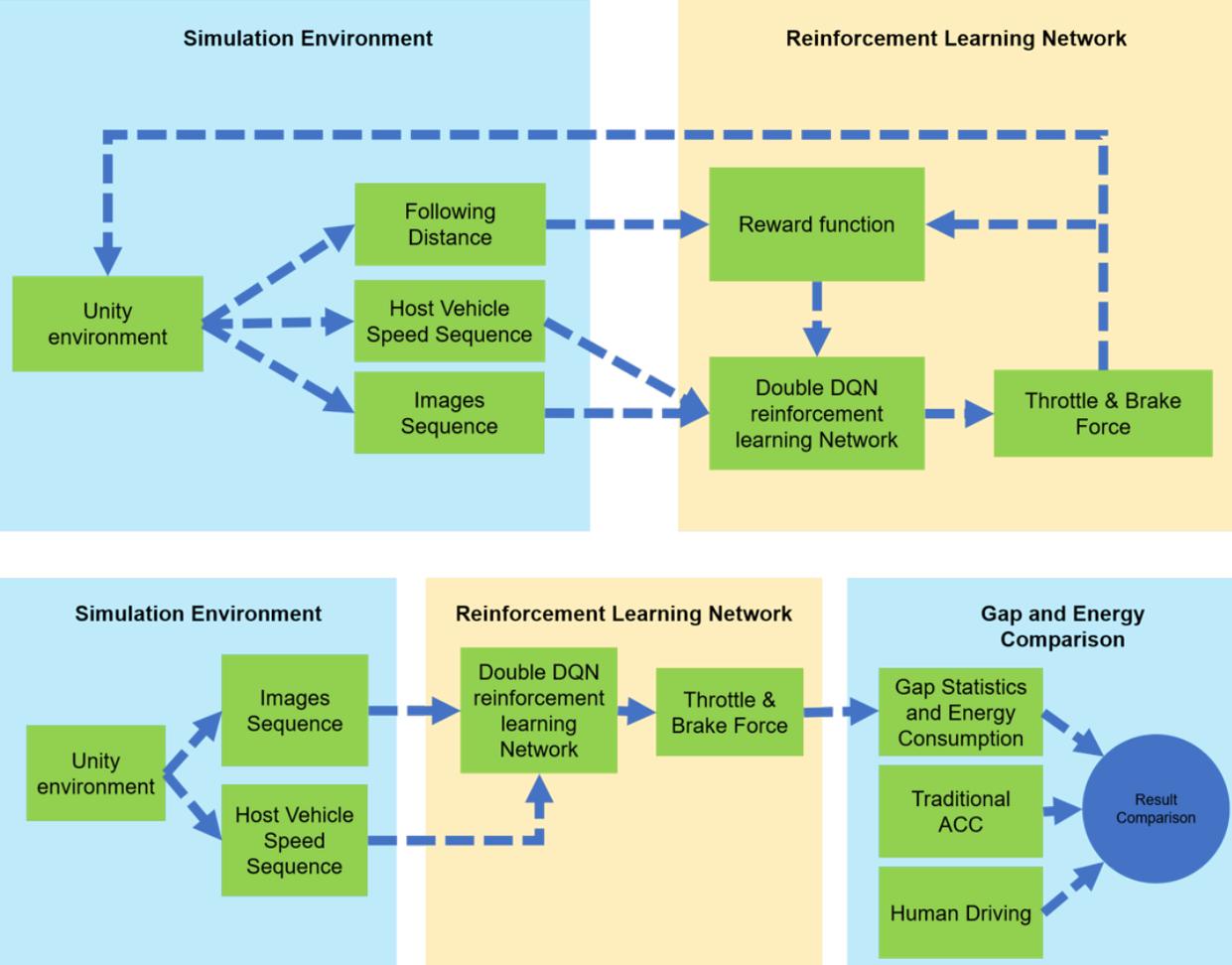

**Figure 1 System Workflows:** System workflow for training (upper) and testing (lower).

**Experiment Setup**

In order to train and test our algorithm, Unity, a well-developed game engine was used for the simulation study. We built a freeway scene with trees, buildings, and traffic flows for our simulation environment, as shown in **Figure 2**. Shadows of all the objects in the environment can be changed with the global light. There are three lanes in each direction along the two-way freeway segment. The preceding vehicle and host vehicle are on the middle lane, while faster traffic flow is on the left lane, and slower traffic flow is on the right lane. The host vehicle is designed to always follow the same preceding vehicle, while the lane change maneuver is not allowed. All the objects can be controlled by Unity's C# scripting Application Programming Interface (API) to reach the desired states. Specifically, the simulation we customized mainly provides the following four functions:

1. *Information Communication*: The information, such as velocity, following distance, and images, is transmitted to the reinforcement learning (RL) network by a socket API using User Datagram Protocol (UDP). The suggested throttle/brake force is calculated and transmitted back to the simulation environment using the same API.

2. *Host Vehicle Control*: The host vehicle can be controlled by the suggested throttle/brake force from the RL network. It also can be controlled by other methods, such as the traditional ACC algorithm, or the human-in-the-loop simulation.

3. *Preceding Vehicle Control*: The velocity data of the preceding vehicle is generated from a large pool of trajectories. Both virtual trajectories and real-world data are used during training and testing.

4. *Simulation Environment Reset*: When a collision occurs or the vehicle following distance is greater than 300m, the current episode is terminated, and the simulation environment is reset with all the object settings to be new initial states for the next episode.

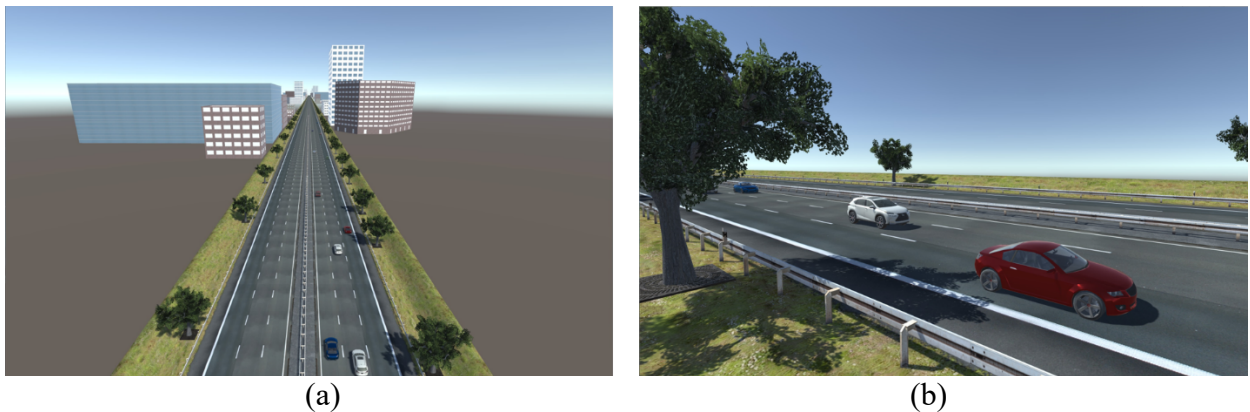

(a)                  (b)

**Figure 2 Simulation Environment.** Bird's eye view (a) and elevation view (b) of the simulation environment.

**Data Preprocessing**

As shown in **Figure 3**, the captured image of size 400 by 400 is firstly transferred to grayscale and both its height and width are resized by 0.5. Then a mask is applied to each captured image, so only the useful information of the preceding vehicle is kept. Finally, we crop out the top border, so the smaller image of size 105 by 200 has less noise and can save computational power for both training and testing.

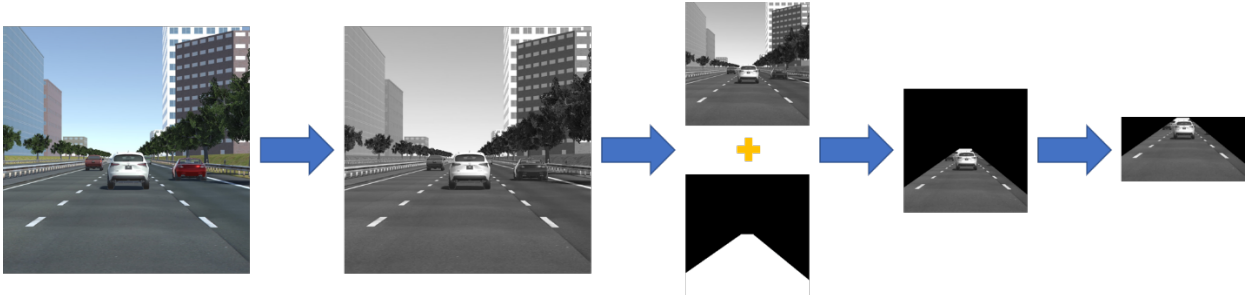

**Figure 3 Image Preprocessing Steps.** The captured image is transferred to grayscale, resized, and cropped after applying the mask.

**DDQN and Network Structure**

RL learns about a certain policy (*π*) to take suitable action (*a*) to maximize the preset reward function given the current state (*s*) of the system with regular communication between the environment and the agent. At each time step *t*, the agent observes the state $s_t$ and takes an action at, which will be implemented back to the environment. The environment receives the action and transfers to a new state $s_{t+1}$. A reward is calculated based on each transition and the total expected reward *R* that we want to maximize is calculated as:

$$R = \sum_{t=1}^{\infty} \gamma^{t-1} * r_t \qquad (7)$$

where *γ* is discount factor ranging from (0, 1). *γ* guarantees the convergence of the optimal control policy and takes both the immediate reward and future reward into consideration.

Based on the RL algorithm, DRL is developed by combining the deep learning (DL) and RL so that the action can be directly computed given a multi-dimensional state, such as a sequence of images based on time series. Deep Q Network (DQN) uses Q Learning as the RL algorithm and convolutional neural network (CNN) as the DL algorithm. CNN receives the given image input and outputs the action value (Q value) corresponding to each possible action. The action with the maximum Q value is chosen according to the CNN output and gets executed to the agent. DQN uses the same Q function for both evaluations and selection of the optimal action, and the problem of overestimating occurs frequently. DDQN avoids this problem by using two Q functions for evaluating and selecting the optimal actions, respectively, thus chosen to be the RL algorithm for this work. The CNN architecture applied in the learning is shown in **Figure 4**.

In the network, the image input is a time series sequence of 8 images with size 105 by 150 in grayscale, and the speed input is a time series sequence of 8 speed values of the host vehicle. *Conv* represents the convolutional layer and *fc* is the fully connected layer. The 21 output values indicate the 21 Q values of the discretized brake/throttle forces to the vehicle at the given state. The reward functions are defined in **Table 1**. Gap is defined as the distance between the leading and following vehicles, and force is discretized and varies from -1 to 1 with negative indicating brake and positive indicating acceleration. The normalized force can lead the vehicle to an acceleration range from -5.5m/s$^2$ to 3.5m/s$^2$. We test two different reward functions for ICE vehicles and EVs, one is only related to the gap (gap-based DDQN), while the other one is also related to force (force-based DDQN). As we can see from the table, we encourage a gap between 30 to 80m and forces from -0.3 to 0.3. The sum of the reward is normalized between -1 to 1 to ensure faster learning.

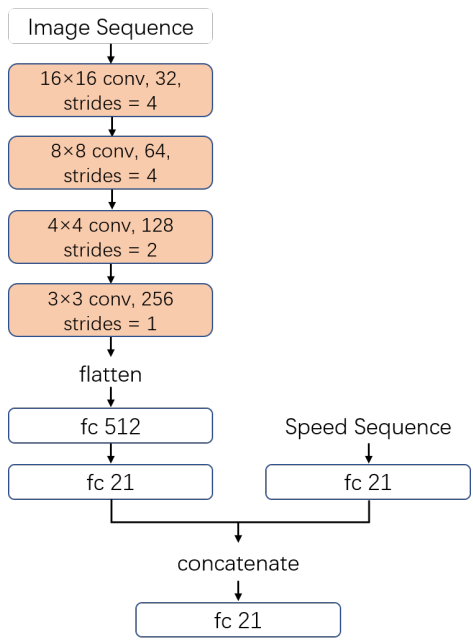

**Figure 4 CNN Structure Applied in the System**

The DRL network is implemented in TensorFlow. The inference time is 3.58 ms. The training, validation, and testing of the network are performed on a PC with six-core 3.70 GHz CPU, 32GB of RAM, and Nvidia GeForce GTX 1080Ti GPU.

**Table 1. Reward Function for Gap-based and Force-based DDQN Model**

| Reward Type | Gap-based DDQN | Force-based DDQN |
|---|---|---|
| **Throttle/Brake Force** | None | *(triangular reward peaking at Force = 0, reward ≈ 10, zero outside [-0.3, 0.3])* |
| **Gap** | *(reward vs Gap: rises from -10 at Gap=0 to peak ≈10 near Gap≈50, then decreases linearly to about -5 at Gap=300)* | |

**CASE STUDY AND RESULTS**

As aforementioned, the preceding vehicle trajectories are either generated virtually or collected from the real world. For simulations of ICE vehicles, trajectories of the leading vehicle are virtually created by an accumulation of 10 sinusoidal functions, shown as below:

$$v_{lead} = 30 + 0.3 \sum_{i=1}^{10} A_i \sin(2\omega_i t) \tag{7}$$

where $A_i$ ($i = 1,2,...,10$) are random integer values of 1 or -1, $\omega_i$ ($i = 1,2,...,10$) are random numbers between 0 and 1, and $t$ is the elapsed time in the simulation environment with a resolution of 0.02 sec. During the training, the group of random 10 $A_i$s and 10 $\omega_i$s will be updated when a new episode begins to make sure that the trajectories are different for each episode. During the testing, a preset 4 groups of $A_i$s and $\omega_i$s are used to generate the trajectory for comparison.

For simulations of EVs, more than 100 hours of electric vehicle driving data are collected under real-world conditions using a 2012 Nissan Leaf (22). The data from a freeway segment (SR-91 in Riverside, California, USA) are applied to the training and testing.

Depending on the different scenarios for ICE vehicles or EVs in the simulation, the parameters of preceding vehicle (PV) and host vehicle (HV) are shown in **Table 2**.

**Table 2. Trajectory Parameters for ICE Vehicles and EVs in the Simulation**

| Parameters | ICE Vehicle | Electric Vehicle |
|---|---|---|
| **PV velocity range (m/s)** | [27, 33] | [0, 30] |
| **PV acceleration range (m/s²)** | [-3.5, 3.5] | [-5.5, 3.5] |
| **Initial PV velocity range (m/s)** | [27, 33] | [11.2, 15.6] |
| **Initial HV velocity range (m/s)** | [25, 30] | [11, 16] |
| **Initial following distance range (m)** | [25, 35] | [25, 35] |

For testing, four trajectories with 4 minutes each are generated to compare between different methods. The training and testing results of the proposed end-to-end vision-based ACC models (with different reward functions), traditional ACC model, and human-in-the-loop simulation are shown in the next two subsections. Note the human-in-the-loop simulation is conducted by 3 different drivers on a driving simulator platform, which is built by on a Windows desktop (processor Intel Core i7-7700K @ 4.20GHz, memory 64.0GB), Unity (version 2018.3.12f), and a Logitech G27 Racing Wheel (model W-U0001)

**Application to ICE Vehicle**

For the gap-based DDQN model, the designed reward function encourages a stricter follow (in terms of gap) to the preceding vehicle regardless of the force used in the process. The average reward with respect to time steps in the training process is shown in **Figure 5.a**. As we can see from the graph, the reward is gradually increasing to 1, meaning that the host vehicle has learned to keep a good distance to the preceding vehicle. For the force-based DDQN model, the reward function encourages not only a good following to the preceding vehicle but also least force (either tractive or brake force) used in the process. The average reward with respect to time steps in the training process is shown in **Figure 5.b**. Compared to the previous model, the training takes longer to converge due to more complex factors in reward functions.

The following distances of gap-based DDQN, force-based DDQN, traditional ACC, and human-in-the-loop models during testing are shown in **Figure 5.c**. The 55m desired distance which leads to the highest gap reward is also labeled in the graph. As we can see from the graph, the gap-based DDQN model follows the desired gap most precisely, which fulfills the target of the model. The force-based DDQN model performs the worst in terms of the gap keeping ability due to the force saving objective.

The trajectories of the host vehicle following the same leading vehicle in all the four methods are shown in **Figure 5.d**. Compared to our force-based DDQN model, the host vehicle of the gap-based DDQN model follows the preceding vehicle more strictly, thus more sensitive to speed fluctuation of the predecessor. This may result in higher fuel consumption. In addition, the use of throttle/brake force as the reward function may lead to the pulse and glide (PnG) driving style which is also considered an eco-driving strategy for ICE vehicles (23).

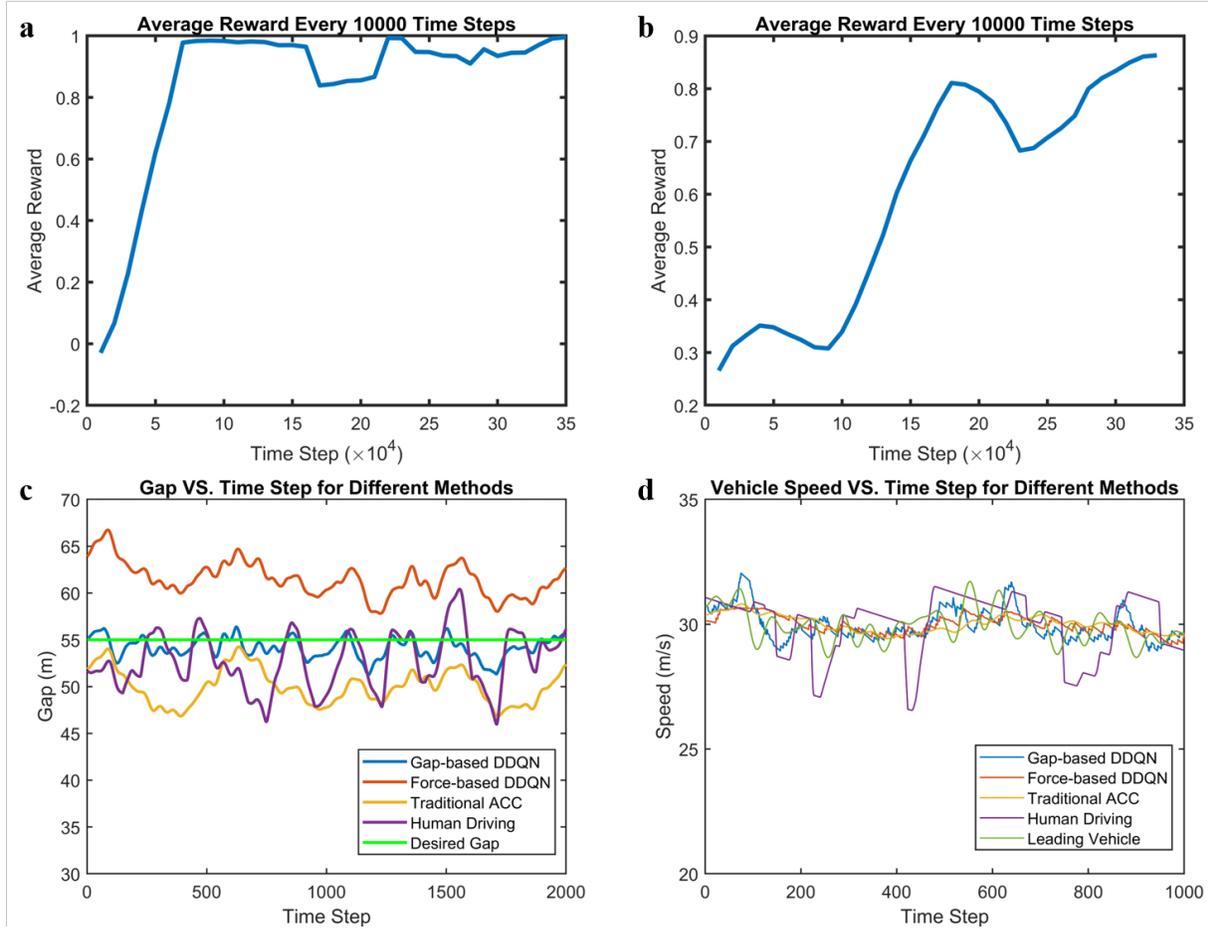

**Figure 5 Training and Testing Result for Virtual Trajectory.**

The energy consumption, pollutant emission, and gap root mean square error (RMSE) of all methods are shown in **Table 3**.

**Table 3. Energy Consumption and Pollutant Emission for Combustion Engine Vehicles between different methods**

| Method | Leading Vehicle | Gap-based DDQN | Force-based DDQN | Traditional ACC | Human Driving |
|---|---|---|---|---|---|
| **Fuel Rate (g/mi)** | 195.393 | 169.275 | 143.754 | 143.805 | 198.586 |
| **$CO_2$ (g/mi)** | 346.352 | 339.069 | 330.681 | 337.124 | 368.524 |
| **CO (g/mi)** | 143.214 | 99.003 | 58.119 | 55.433 | 143.391 |
| **HC (g/mi)** | 13.613 | 12.913 | 9.843 | 9.443 | 10.673 |
| **$NO_x$ (g/mi)** | 4.656 | 3.159 | 1.841 | 1.743 | 4.500 |
| **Gap RMSE (m)** | N/A | 3.423 | 7.686 | 6.667 | 6.591 |

**Application to EV**

For the EV model, we use the same reward function as ICE model. The average reward, gap, and sample trajectory with respect to time steps in the training and testing process are shown in **Figure 6**, which has the same layout as **Figure 5**. The energy consumption and gap RMSE of all methods are shown in **Table 4**.

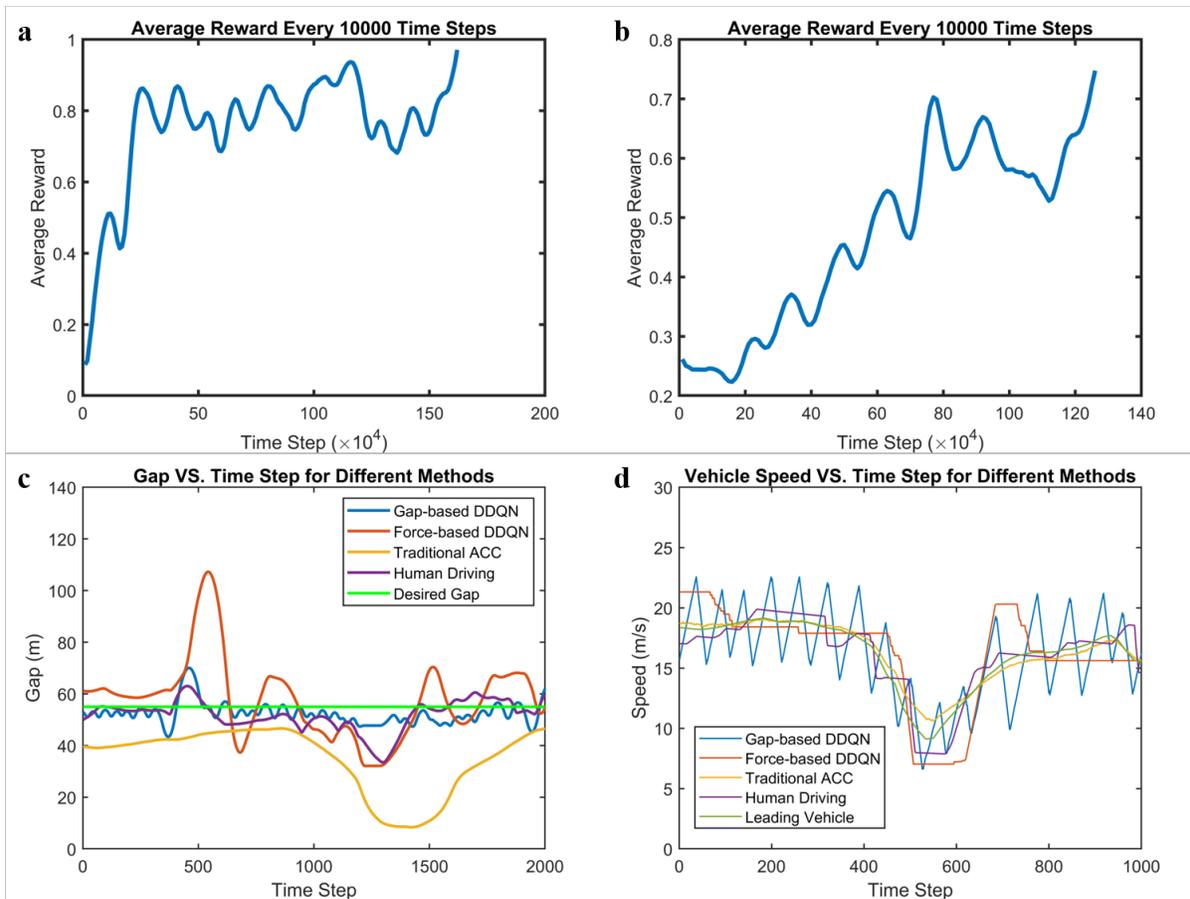

**Figure 6 Training and Testing Result for Real Trajectory.** The figure layout for EV model is the same as the ICE model result layout in **Figure 5**.

Table 4. Energy Consumption for EV between different methods

| Method | Leading Vehicle | Gap-based DDQN | Force-based DDQN | Traditional ACC | Human Driving |
|---|---|---|---|---|---|
| **Energy Rate (KJ/mile)** | 1.135e3 | 4.914e3 | 2.056e3 | 1.128e3 | 1.500e3 |
| **Gap RMSE (m)** | N/A | 5.083 | 10.98 | 17.440 | 5.439 |

The same trend of training convergence in the ICE model can be observed in the EV model. The training takes longer to converge in the force-based DDQN model due to more factors in reward functions. The gap-based DDQN model results in the best gap regulation among all the methods, which is the same as in the case of ICE vehicle application. Due to the more abrupt and frequent speed fluctuations, such a strict gap regulation strategy results in the highest EV energy consumption among all the methods (much higher than others). Unlike the application in ICE vehicle, the force-based DDQN model does not result in better energy consumption for EV compared to the performance of human-in-the-loop simulation or traditional ACC, although the PnG-like behavior is conducted (see **Figure 6.d**). As expected, the gap is quite loosely regulated by the force-based DDQN model.

**CONCLUSIONS**

This research explored an end-to-end vision-based ACC using DRL. We constructed the Unity game simulation environment and used socket connection to communicate with reinforcement learning network. We designed gap-based and force-based reward functions in training ICE or EV models. Training and testing result over virtually generated and real-world driving trajectories show the effectiveness and robustness of the following ability. Compared to traditional ACC and human-in-the-loop simulation, the gap-based following system can achieve the best following ability but sacrifice energy saving performance. For ICE model, the force-based following system only achieves a better energy saving performance tested on virtually generated trajectories. The inference time of 3.58 ms indicates the real-time working ability of the proposed method.

In the future, more research will be conducted as listed below:
• Integrate the energy model into DRL reward function for an Eco-ACC system
• Other reinforcement learning algorithms, such as Actor-Critic, which can output continuous force sequences, will be tested for comparison.
• Different recurrent neural network (RNN) structures, such as Long short-term memory (LSTM), will be tested with the entire sequence of data, and compare with the current network result.

**ACKNOWLEDGMENTS**
We would like to thank Mr. Chao Wang and Dr. Fei Ye from Center for Environmental Research and Technology for their valuable inputs.